\documentclass[11pt]{article}

\usepackage{acl}

\usepackage{times}
\usepackage{latexsym}
\usepackage{graphicx}
\usepackage{multirow}
\usepackage{booktabs}
\usepackage[T1]{fontenc}

\usepackage[utf8]{inputenc}

\usepackage{microtype}

\title{A Robustly Optimized Long Text to Math Models for \\ Numerical Reasoning On FinQA}

\author{Renhui Zhang \and Youwei Zhang \and Yao Yu \\
zhangrenhui.zrh@antgroup.com \\ zhangyouwei.zyw@antgroup.com \\
csyuyao@gmail.com
}

\begin{document}
\maketitle
\begin{abstract}
Numerical reasoning is required when solving most problems in our life, 
but it has been neglected in previous artificial intelligence researches. 
FinQA challenge has been organized to strengthen the study on numerical reasoning where the participants are asked to predict the numerical reasoning program to solve financial question.
The result of FinQA will be evaluated by both execution accuracy and program accuracy.
In this paper, we present our approach to tackle the task objective by developing models with different specialized capabilities and fusing their strength. 
Overall, our approach achieves the \textbf{1st place} in FinQA challenge, with \textbf{71.93\%} execution accuracy and \textbf{67.03\% }program accuracy. 
\end{abstract}

\section{Introduction}
Numerical reasoning is an useful and important ability of human being, which is also one goal of artificial intelligence.
However, most researches focus on obtaining the right answer, ignore the reasoning ability of models \citep{talmor2018web, yang2018hotpotqa}.
In 2019, the DROP dataset is introduced to study the numerical reasoning ability of deep learning models \citep{dua2019drop}.
The DROP dataset contains only simple numerical reasoning problems, which are far from practical problems.

In order to improve the numerical reasoning ability of artificial intelligence, Chen introduce the FinQA dataset \citep{chen2021finqa}.
FinQA contains 8,281 financial question-answer pairs and the numerical reasoning program, which are annotated by experts.
The task of FinQA is predicting the numerical reasoning program, based on the supporting facts that retrieved from the financial report.
Different to previous dataset, FinQA evaluates the result by two metrics, program accuracy and execution accuracy.
Program correct means that the predicted program is logically identical to the ground truth.
And execution correct means that the answer obtained by executing the predicted program is equal to the ground truth.

Our team first mitigates the data distribution gap between training and test set by adversarial validation.
Then we develop models that have different specialized capabilities and fuse their strengths to prevent over-fitting and obtain higher performance.
Our method achieves 71.93\% execution accuracy and 67.03\% program accuracy, which is the 1st place in FinQA challenge. 
In this paper, we mainly present our approach, with more details about the architecture including experimental studies. 
We also provide our code to the community, which will be freely available to everyone interests in this work\footnote{The GitHub repository is publicly available on: \url{https://github.com/UVformal/LTMM}}.

\begin{figure}[t]
    \centering
    \includegraphics[width=\linewidth]{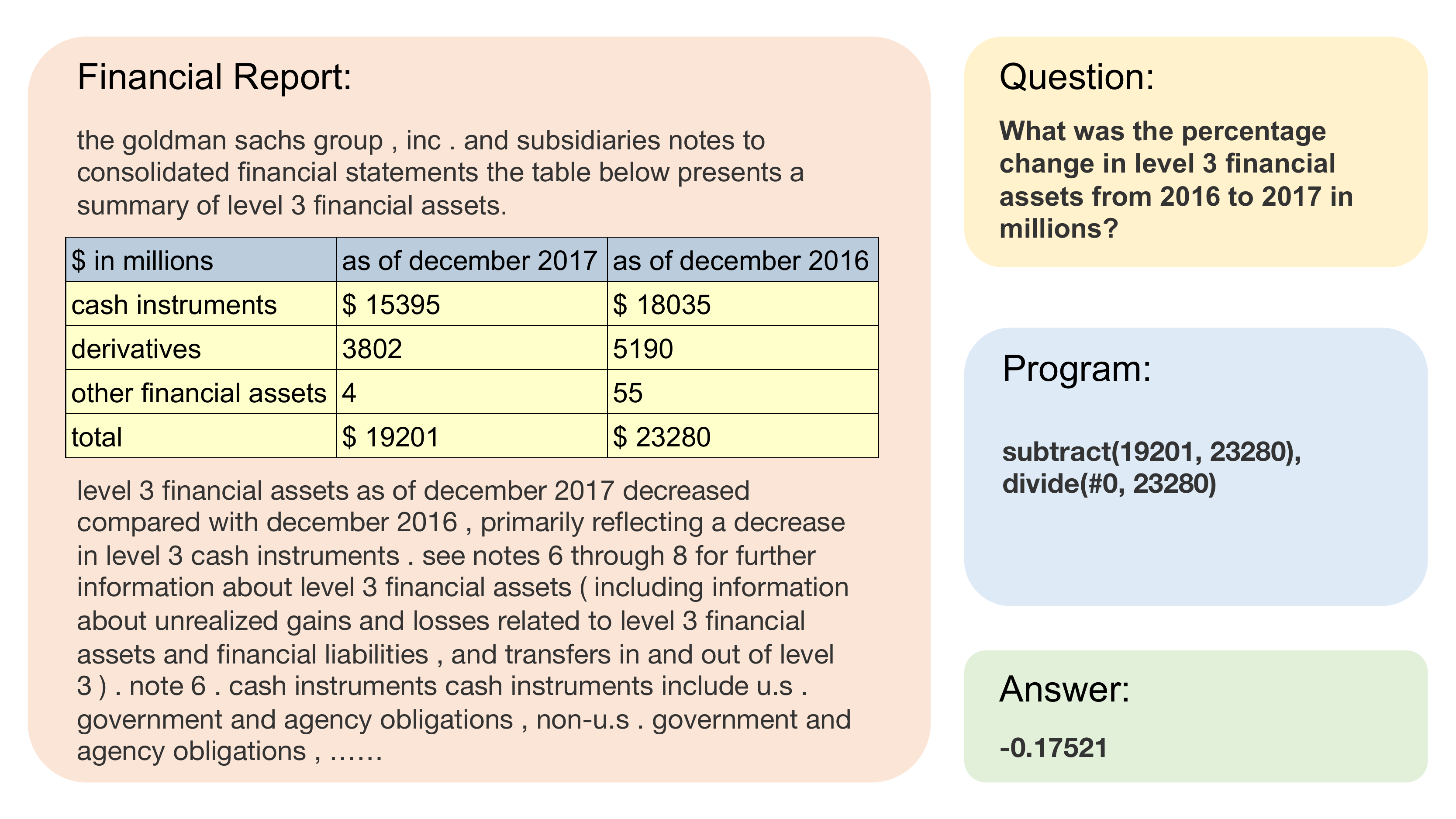}
    \caption{An example in FinQA dataset, which consists of question, financial report, reasoning program and the answer.}
    \label{fig:example}
\end{figure}

\section{Related Work}
\quad\textbf{Adversarial Validation.} 
Adversarial validation \citep{AdversarialValidation} is a method to deal with the data distribution gap between training and test set.
The core idea is building a classifier to distinguish whether the example comes from the training or the test set.
If the classifier cannot distinguish the source of the sample, it means that the data distribution of training and the test set are relatively consistent; 
otherwise, the data distribution of the training set and the test set are not consistent.
With adversarial validation, we can build a dataset which has relatively consistent distribution with the test set.

\textbf{Pre-trained Language Model.} 
Since Transformer is proposed \citep{vaswani2017attention}, Pre-trained Language Models (PLM) have achieved great success in various fields of Natural Language Processing (NLP).
Bert \citep{devlin2018bert} is the first large-scale pre-trained language representation model. 
It uses masked language model (MLM) to pre-train bidirectional transformers to generate deep bidirectional language representations.
After pre-training, it only needs to add an output layer for fine-tuning to achieve state-of-the-art performance in a variety of downstream tasks.
Roberta \citep{liu2019roberta} is an improved version of Bert, it achieves better performance by pre-training with larger model, larger bacth size and more training data.
T5 \citep{raffel2019exploring} converts tasks such as translation, classification, regression, and summary generation into Text-to-Text tasks, thus these tasks can use the same objective function during training (pre-train and fine-tune) and testing. 
Thus the pre-trained information can be efficiently transferred to downstream tasks.

\textbf{Ensemble Learning.}
Ensemble learning \citep{polikar2012ensemble} is a process of fusing multiple models according to a certain method. 
Ensemble learning is mainly used to improve the performance of models or to reduce the possibility of selecting improper model. 
The ensemble algorithm itself can be a supervised learning algorithm because it can be trained and then predicted.

\section{Data}
\subsection{FinQA Dataset}
The FinQA dataset\footnote{The dataset can be downloaded from \url{https://github.com/czyssrs/FinQA}} 
contains 6251, 883 and 1147 examples in training, validation and test set respectively.
As shown in Figure \ref{fig:example}, the example in FinQA dataset consists of question, financial report, reasoning program and the answer.
Each financial report contains amounts of facts, 
unstructured texts and structured tables, and tables has been converted to text.

The task of FinQA is first understanding the question and retrieving the supporting fact from financial report, then predicting the numerical reasoning program and obtaining the answer.
Each operation is represented by four steps in the reasoning program, for example, the operation "add(a,b)" is represented by ["add(", "a", "b", ")"] in the reasoning program.
The first step is operator, such as "add(", "subtract(";
the 2nd and 3rd steps are numbers, which can be numbers from the support facts, constant number or the result of previous operation that saved by step memory;
the 4th step is fixed to ")".
Besides, an end sign "EOF" will be added to the end of the reasoning program.
By executing the reasoning program, we can obtain the execution answer.
Finally, the result will be evaluated by both program accuracy and execution accuracy.

\subsection{Distribution Gap}
We applied adversarial validation \citep{AdversarialValidation} to analyze the data distribution of FinQA.
All examples from training, validation and test set are mixed together, and a classifier is trained to distinguish the source of each example.
The classifier achieves 0.91 AUC during training, and 0.85 AUC during testing.
It means that there is a great distribution gap between the test set and others.

By re-splitting the training and validation set, we construct a new validation set, which contains 900 examples, to reduce the distribution gap between validation and test set. 

\begin{table*}[]
\centering
\caption{The experimental results of retriever on validation and public test set without adversarial validation.}
\label{tab:retriever}
\begin{tabular}{l|cc|cc}
\toprule
\multirow{2}{*}{Model} & \multicolumn{2}{c|}{Valid Recall (\%)} & \multicolumn{2}{c}{Public Test Recall (\%)} \\ \cline{2-5} 
                       & top-3            & top-5            & top-3            & top-5           \\
                       \midrule
Baseline Model               & 90.87           & 93.97           & 89.80            & 93.28          \\
Prompt Learning        & 91.54           & 95.11           & 90.16           & 94.12          \\
Contextual Model                & 91.10            & 94.69           & 89.22           & 92.86          \\
Stack               & \textbf{92.63}           & \textbf{95.89}           & \textbf{90.77}           & \textbf{94.33}    \\
\bottomrule
\end{tabular}
\end{table*}
\section{Method}
Following the Retriever-Generator architecture proposed by Chen \citep{chen2021finqa}, 
our method also consists of retriever and generator.
Retriever aims to retrieve the supporting facts from the financial report while generator is designed to generate the numerical reasoning program.

Considering the distribution gap in FinQA, single large model that easily to over-fit to training set is not a ideal solution.
Therefore, we develop several small models and fuse them.
Compared with single large model, small models are less prone to over-fit.
Besides, each model has its strength, and we can fuse the strengths of different models to achieve higher performance than single model. 
In this section, we will introduce the proposed method in detail.
\subsection{Retriever}
The financial reports in FinQA usually contain amounts of facts, but most of them are not relevant to the question.
The retriever is designed to retrieve the supporting facts that are relevant to the question.

\textbf{Baseline Model.} Following Chen \citep{chen2021finqa}, the baseline model is a binary classification model with a Bert backbone to extract embedding. 
Each fact in the financial report is concatenated with the question, and a classifier is trained to distinguish whether the fact is supporting fact.
At inference stage, all facts are ranked by the prediction score and the top-$k$ are chosen as the supporting facts.

\textbf{Prompt Learning.}
\begin{figure}
    \centering
    \includegraphics[width=\linewidth]{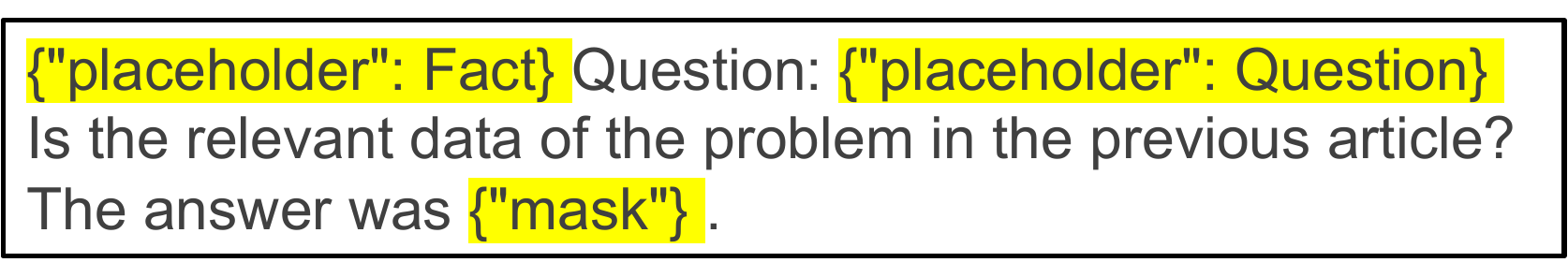}
    \caption{Template of prompt learning.}
    \label{fig:prompt_template}
\end{figure}
Prompt learning \citep{liu2021pre} is an improved pre-training scheme, where the pre-trained information can be efficiently transferred to downstream tasks.
It works significantly when the data of downstream task is insufficient, which is suitable to FinQA.
As shown in Figure \ref{fig:prompt_template}, there is a template of prompt learning, we need to replace the placeholder with specific fact and question.
The prediction, "yes" or "no", will be outputted at the position of "mask". 
Same to baseline model, we choose the top-$k$ facts according to the prediction score.

\textbf{Contextual Model.}
The above two models processed one fact each time without contextual information.
In order to make full use of the contextual information, a contextual model is proposed.
This model concatenates question and $n$ facts and predicts these $n$ facts together.
Unlike the baseline model, contextual model predicts whether the fact is the start/end position of the supporting facts, which is same as the pre-trained task of the backbone Roberta\citep{liu2019roberta}.
Besides, sliding window with step $[n/2]$ is applied to make sure that model predict each fact with the contextual information.
At the inference stage, all facts are ranked by the sum of start score and end score and the top-$k$ are chosen.
\begin{table}[]
    \centering
    \caption{Comparison between Stack and Stack Positive on public test set.}
    \label{tab:ensemble_positive}
    \resizebox{\linewidth}{!}{
    \begin{tabular}{l|cc|cc}
    \toprule
\multirow{2}{*}{Model} & \multicolumn{2}{c|}{Recall (\%)} & \multicolumn{2}{c}{Precision (\%)} \\ \cline{2-5} 
                       & top-3             & top-5             & top-3              & top-5              \\ \midrule
Stack                  & \textbf{90.77}             & \textbf{94.33}             & 47.57              & 30.20              \\
Stack Positive         & 90.42             & 93.21             & \textbf{55.22}              & \textbf{43.36}       \\
\bottomrule
\end{tabular}}
\end{table}

\textbf{Model Fuse by Stacking.}
The above three models can achieve high retriever performance, and they are expert in retrieving different samples.
Therefore, we can fuse their strengths to obtain higher performance than single model, by training a logistic regression classifier to ensemble the predictions of these three models.
Ensemble can reduce the possibility of selecting improper model, and improve the retriever performance.
Practically, we feed the prediction score of these three models to the logistic regression classifier and obtain a new score.
Finally, we rank all facts by the new score and choose the top-$k$ as the supporting facts.

\subsection{Generator}
Generator usually consist two parts, encoder and decoder.
Encoder take the question and supporting facts as input and encode them to embedding, decoder take the embedding as input and decode the numerical reasoning program step by step.

\begin{table*}[h]
    \centering
    \caption{The experimental results of generator on validation and private test set.}
    \label{tab:generator}
\begin{tabular}{l|cc|cc}
\toprule
\multirow{2}{*}{Model} & \multicolumn{2}{c|}{Valid Accuracy (\%)} & \multicolumn{2}{c}{Private Test Accuracy (\%)} \\ \cline{2-5} 
                       & exe acc            & prog acc            & exe acc           & prog acc           \\ \midrule
Baseline Model         & 68.78              & 66.22               & 67.79             & 63.66            \\
Remove Redundancy      & 69.72              & 66.89               & 66.70             & 62.67              \\
Transformer Decoder    & 70.55              & 67.27               & 67.46             & 63.43              \\
Vote                 &       \textbf{73.67}             &    \textbf{70.52}                & \textbf{71.93}            & \textbf{67.03}        \\
\bottomrule
\end{tabular}
\end{table*}
\textbf{Baseline Model.}
Following Chen \citep{chen2021finqa}, the baseline generator use Bert as encoder and bidirectional LSTM \citep{hochreiter1997long} as decoder.
Besides, Chen add several attention layers between the encoder and decoder to make the model focus on the relevant embedding.
And after decoding an operation, the step memory that stores result of this operation will be updated.
When the output is end sign, "EOF", the decode progress will stop.

\textbf{Remove Redundancy.}
The baseline model feed all embedding, both numbers' and words', to decoder.
However, most words' embedding is helpful less in decoder.
The redundant words' embedding became a burden on the model, increasing the learning difficulty and reducing the efficiency.
Therefore, we design a model that only feed the numbers' embedding to decoder.
This model focus more on the numbers, thus it is more suitable for predicting numbers than the baseline model.

\textbf{Transformer Decoder.}
For the sake of higher performance and efficiency, we replace the decoder in the baseline by a four layers transformer decoder \citep{vaswani2017attention}.
Compared to LSTM, transformer decoder can learn contextual information more sufficient, makes the model focus on the most relevant information to the question.
Therefore transformer decoder can achieve higher performance than LSTM.
Besides, transformer decoder can be trained in parallel, which is more efficient.

\textbf{Model Fuse By Voting.}
Same as retriever, we also fuse the above three models by voting to fuse the strengths of different models.
We assign the execution accuracy of each model as the weight of voting progress.
And the program with most votes will be selected as the final result.

\section{Experiments}
\subsection{Implementation Details}
Our experiments in this paper are conducted with 4 Tesla-V100 GPUs.
In order to stabilize the training process and prevent over-fitting, we adopt the following schemes:
\begin{itemize}
    \item Freezing the parameter of the backbone, which is a pre-trained model, at the first 10 epochs;
    \item Scheduling the learning rate to warm up at beginning and  gradually decrease the learning rate during training;
    \item Clipping the gradient prevents gradient explosion;
    \item Applying weight decay to prevent over-fitting.
\end{itemize}
Besides, we set $n=8$ for contextual model and $k=3$ for all retriever.

\subsection{Results of Retriever}
Extensive studies on retriever with varied models are conducted and the results are shown in Table \ref{tab:retriever}.
It can be seen that these three single models both achieve high top-3 recall performance on the test set.
And by fusing these three models, we further improve 1.09\% and 0.61\% top-3 recall performance on validation and test set respectively.

However, the average number of supporting fact in FinQA is 1.71 \citep{chen2021finqa}, which means that many retrieved supporting facts are false positive.
Therefore, we propose Stack Positive, which selects only positive facts in top-$k$ as the supporting facts. 
The comparison between Stack and Stack Positive is shown in Table \ref{tab:ensemble_positive}.
Although the recall performance of Stack Positive is slightly lower than Stack, 
Stack Positive achieves higher precision.
In other words, Stack Positive can provide more precise supporting facts to generator than Stack and other models.

\subsection{Results of Generator}
Based on the supporting facts provided by retriever, 
we conduct experiments on generator and the results are shown in Table \ref{tab:generator}.
It can be seen that every single model can achieve high performance on validation set, but performs poorly on test set.
To counter the distribution gap, we ensemble these three models by voting and fuse their strengths.
By fusing, we can get the best performance, which has a improvement of 4.14\% execution accuracy and 3.37\% program accuracy over single model on private test.

\subsection{Ablation Studies}

Table \ref{tab:adversarial_validation} shows the ablation study results of adversarial validation. 
We train same model with same settings the original dataset and re-split dataset.
We select checkpoints according to the performance on validation dataset, and test them on the private test set.
It can be seen that we can choose a checkpoint that is suitable to private test set on re-split dataset than original dataset.
Compared with model trained by original dataset, model trained by re-split dataset performs better on test set, and the performance gap between validation and test is reduced.
\begin{table}[]
    \centering
    \caption{Ablation study of adversarial validation on validation and private test set.}
    \label{tab:adversarial_validation}
    \resizebox{\linewidth}{!}{
    \begin{tabular}{l|cc|cc}
    \toprule
\multirow{2}{*}{Dataset} & \multicolumn{2}{c|}{Valid Accuracy (\%)} & \multicolumn{2}{c}{Test Accuracy (\%)} \\ \cline{2-5} 
                         & exe acc            & prog acc            & exe acc           & prog acc           \\ \midrule
Original                 & \textbf{69.54}            & \textbf{66.59}             & 66.81             & 62.24              \\
Re-Split                 & 68.78              & 66.22               & \textbf{67.79}             & \textbf{63.66}      \\ \bottomrule       
\end{tabular}}
\end{table}


\begin{figure}[t]
    \centering
    \includegraphics[width=\linewidth]{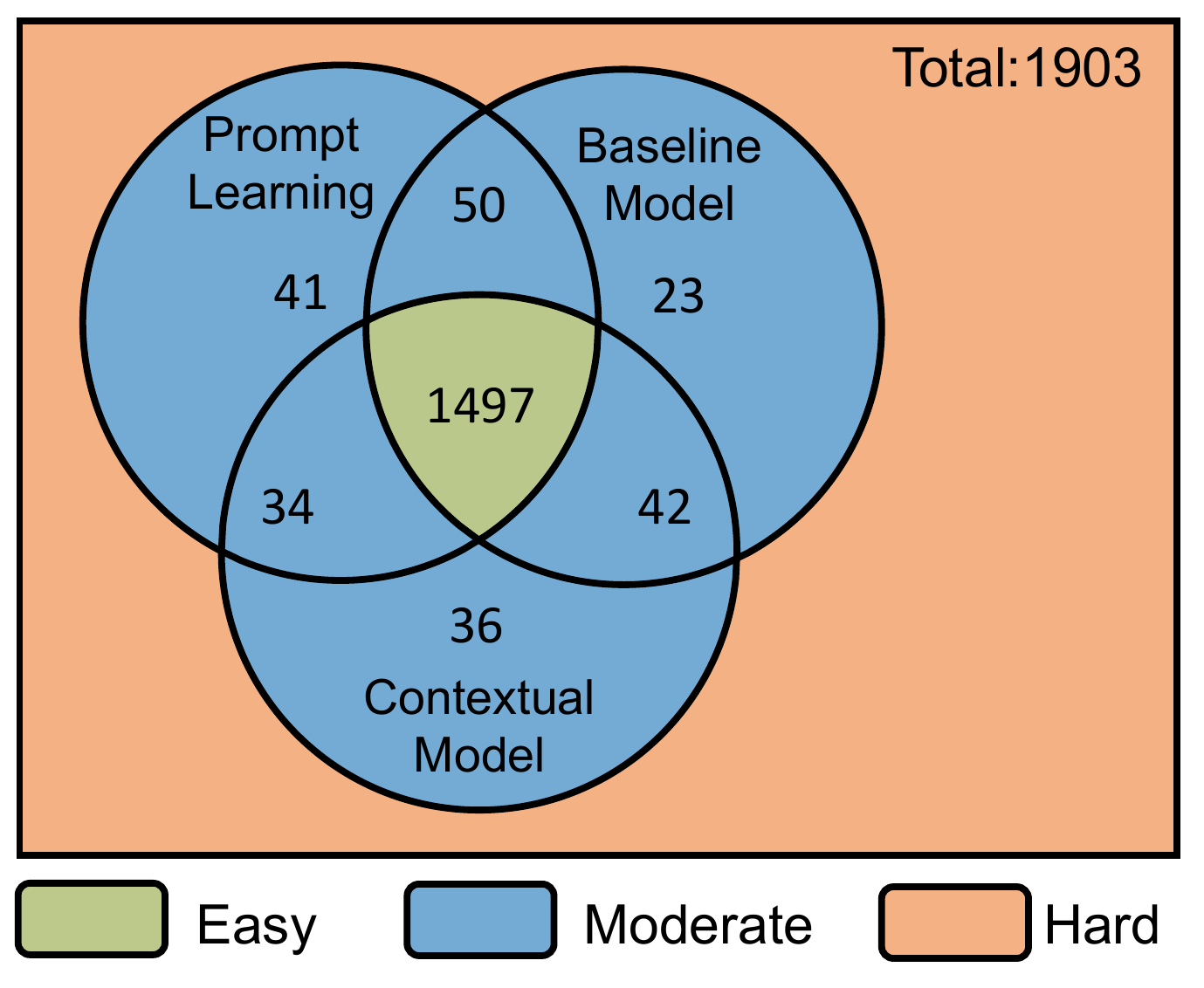}
    \caption{The venn diagram of the retrieved facts of three retriever models. Different colors represent examples of different difficulty.  }
    \label{fig:venn}
\end{figure}

\begin{figure}[t]
    \centering
    \includegraphics[width=\linewidth]{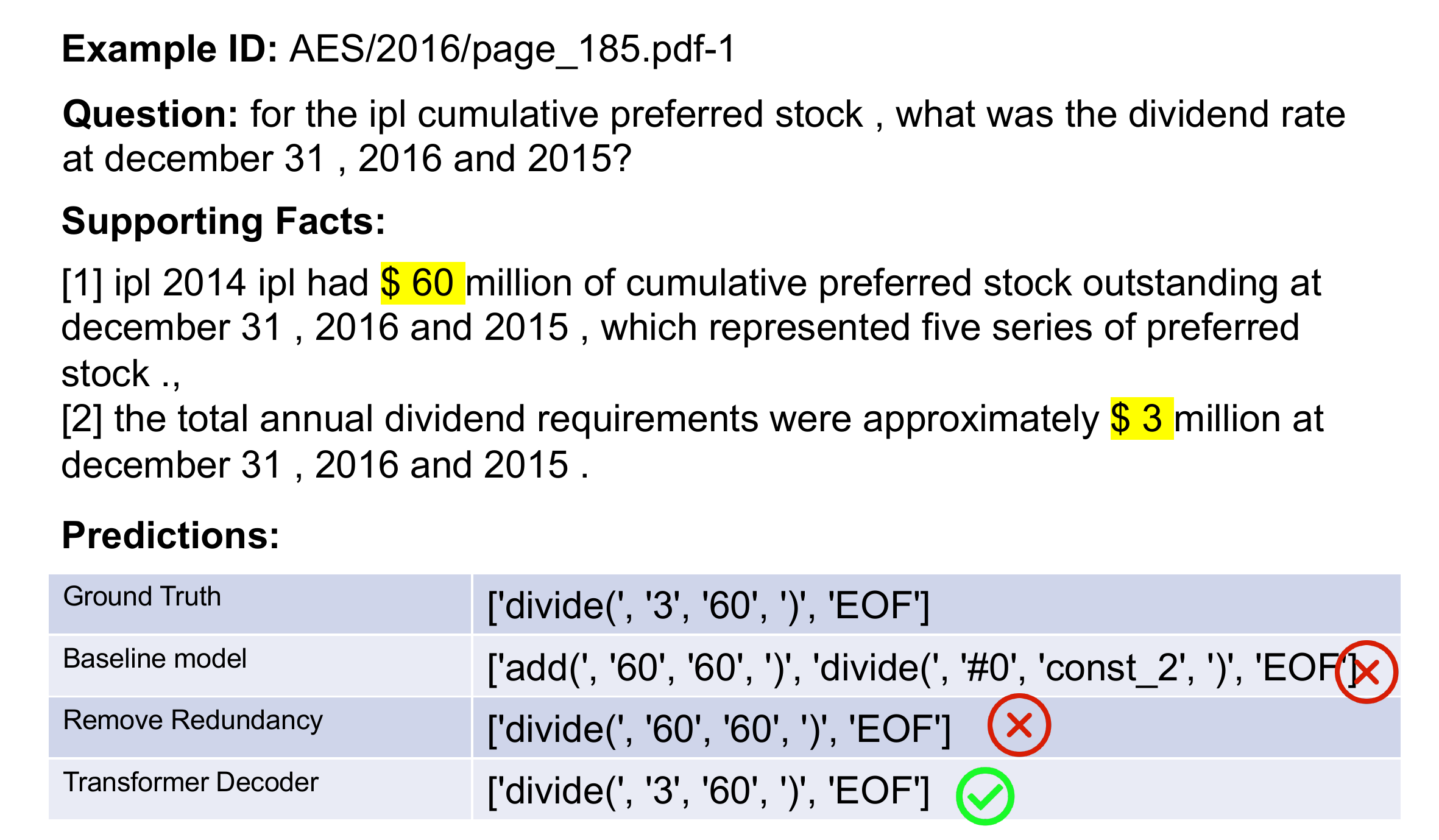}
    \caption{The difference between predictions of different models.}
    \label{fig:ab_retriever}
\end{figure}

Different models have different strengths even they are trained by the same data.
As shown in Figure \ref{fig:venn}, which is the venn diagram of the retrieved facts of three retriever models.
These three circles represent three models, and the overlapping parts means facts retrieved by both models.
It can be seen for easy examples, all these models predict correctly; on the contrary, these models make mistakes on extremely hard examples.
But for examples with moderate difficulty, different models will have different prediction results. 
The difference between models can also be observed in generator, as shown in Figure \ref{fig:ab_retriever}.
Different models predict various numerical reasoning programs, thus we can fuse the right predictions of these models on examples with moderate difficulty to obtain a higher performance.

\section{Conclusion}
In this paper, we propose a method that developing models that have different specialized capabilities and fuse their strengths to tackle the task objective of FinQA.
We first analyze the data distribution of FinQA by adversarial validation, 
and build a new validation dataset that has a similar distribution to test set by re-splitting the training and validation set.
Besides, in order to reduce the influence of distribution gap, we develop and fuse several small models rather than train a large model which is easily over-fitting to training set.
Furthermore, we obtain a high performance by fusing the strengths of different small model.
Overall, we achieve the 1st place in FinQA challenge with 71.93\% execution accuracy and 67.03\% program accuracy.
\section*{Acknowledgements}
We would like to thank the FinQA challenge organizers
for making this constructive dataset available.
\bibliography{anthology,custom}

\appendix



\end{document}